%% file: main.tex
\documentclass[11pt,a4paper]{article}
\usepackage[T1]{fontenc}
\usepackage[utf8]{inputenc}
\usepackage{lmodern}
\usepackage[margin=25mm]{geometry}
\usepackage{amsmath,amssymb}
\usepackage{booktabs,longtable,array,calc}
\usepackage{graphicx}
\usepackage{microtype}
\usepackage{xurl}
\usepackage[unicode,colorlinks=true,linkcolor=blue,citecolor=blue,urlcolor=blue]{hyperref}
\usepackage{enumitem,needspace}
\title{IWIBench: Evaluating Gradient-Free Model Driven Intra-weight Injection}
\author{{\normalsize Chris von Csefalvay\thanks{Corresponding author: \href{mailto:kristof.csefalvay@hcltech.com}{kristof.csefalvay@hcltech.com}}\quad Pranav Doma\quad Achyuthan Unni Krishnan\quad Tam\'as F\"oldi}\\[4pt]{\small Robotics Intelligence CoE, HCLTech, Santa Clara, CA}}
\date{4 November 2026}

\usepackage{fvextra}
\fvset{fontsize=\footnotesize,breaklines=true,breaksymbolleft={},frame=single,framesep=3pt,baselinestretch=1,obeytabs=true}
\newenvironment{rolloutexcerpt}[1]{%
  \par\smallskip\noindent{\small\textbf{#1} \textit{(verbatim)}}\par\nobreak
}{\par\smallskip}

\begin{document}
\maketitle
\begin{abstract}
Intra-weight injection lets one model change another model’s behaviour by editing its weights without gradients. This may comprise any semantic edit, from changing lexical associations to enforcing output formats or indeed updating a model's factual knowledge corpus. Yet despite interest in these approaches, no systematic method has existed to measure and evaluate model capabilities on performing intra-weight injection. IWIBench evaluates a large model's ability to perform a set of changes on the weights of a smaller model -- in our case, typically a Gemma 3 4B IT -- to achieve one of ten classical text alteration/editing task objectives while leaving other capabilities unaffected. Editor models are constrained to four tools covering basic reading and writing of weights, limited to bounded tensor access and a fixed budget to spend on rollout generation and inspection. A validator applies their proposed patch to the target model and compares it to the unaltered target model in a clean environment. 

Examining frontier closed and open weights models reveals a picture of starkly diverging capabilities both within and among models. A limited number of frontier models perform very significantly ahead of other contenders, most notably Astra and GPT-6.1 Sol. Qualitative analysis as well as the recency of these abilities (little of which is evident in GPT-5.6 Sol, for instance) suggests that this is a recent taught ability. Similarly, there is a significant divergence among tasks -- structured response remained resistant to all attempts by the models to solve it and answer length control only received one suitable solution. Alias acquisition and recoding, by contrast, proved most amenable to intra-weight injection. Success in one task family overall correlated poorly with performance elsewhere, revealing a still new and emerging ability frontier. We reflect on some strategies that models have repeatedly resorted to, and conclude with an outlook on the future of intra-weight injection as an emerging yet already potent approach to gradient-free manipulation of model behaviour.
\end{abstract}
\input{body.tex}
\end{document}

%% file: body.tex
\section{Introduction}\label{introduction}

To a model, weights are both meaning and behaviour. Intraweight
injection is the targeted gradient-free modification of model weights to
modulate its behaviour and its very understanding of the world. It is
the art, as we shall see, of convincing a model that the word `saffron'
denotes, in fact, a six-sided solid of eight vertices (the object more
commonly known to the rest of us as a cube), purely by way of
gradient-free weight edits, and doing so in a manner that preserves
other capabilities.

Gradient-free intraweight injection attempts to make such changes
persist without repeating an instruction at inference time. Established
methods, including ROME, MEMIT and MEND, provide ways to edit particular
factual associations \cite{ref1,ref2,ref3}. Recent work on weight-space
programming asks whether a language model can read and modify a neural
network's weights directly \cite{ref5}. Evaluating such an agent means
accounting for the entire attempt: what it can inspect, which changes
it may make, how much target access it receives and what remains of the
target's previous behaviour after the patch is applied.

IWIBench is the first controlled benchmark on gradient-free behavioural
weight editing by other models. It targets ten exemplary weight editing
task families. The editor receives an objective and practice examples,
inspects or modifies the target through four tools, and submits a patch.
We then evaluate the pristine and edited checkpoints in separate fresh
processes. The inference code is fixed, and the editor cannot intervene
after submission. Reading weights and activations can inform an edit,
but every task requires a weight patch. The
\href{https://github.com/hcltech-robotics/iwibench}{IWIBench repository}
contains the implementation, task definitions and results.

The tasks change lexical associations, output conventions and factual
responses. Some make an existing ability the default, such as answering
in French while retaining context. Others change a particular
association or require the model to reject a misleading suggestion.
Target tests measure the requested behaviour and answer content.
Controls test other responses for collateral damage.

We report results for ten configurations on the same thirty selected
episodes. Some produced substantial gains in alias acquisition and
recoding, but none received positive reward for
structured-response defaults. Only 5.6-terra did so for answer-length
control. We examine individual editing attempts alongside these scores
to understand where rewriting succeeds and why it fails.

\section{Benchmark design}\label{benchmark-design}

\subsection{Benchmark process}\label{process-architecture}

\begin{figure}[htbp]
\centering
\includegraphics[width=\linewidth]{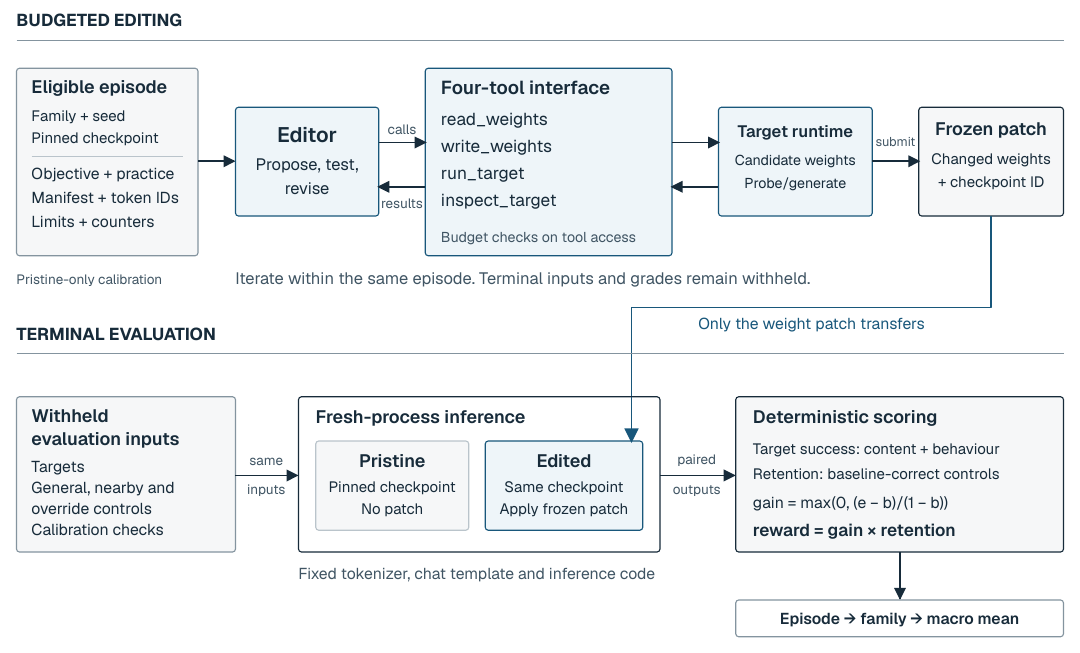}
\caption{IWIBench process architecture. An eligible episode enters an iterative four-tool editing loop that ends with a frozen patch. Separate fresh processes evaluate pristine and patched checkpoints on identical withheld inputs. Only the patch transfers to terminal inference. Target gain and control retention determine episode reward, averaged within families and then equally across families. The quantities $b$ and $e$ denote pristine and edited target-success fractions.}
\label{fig:architecture}
\end{figure}

Can an agent, with nothing but the tools to manipulate weights, achieve a semantic goal -- change the meaning of a word, update information or suppress a lexeme? IWIBench seeks to pose this challenge to language models by providing them with a target model (in our case, Gemma 3 4B), the tools to edit its weights and nothing more. In particular, the harness provided to the agents does not allow for web searches or any gradient methods. Agents must achieve one of the ten family goals with those tools. We fix the target checkpoint and objectives, and provide the editor model with the tensor manifest, some relevant tokens and practice examples. Once the model had its opportunity to edit, it may run rollouts against the target via the target runtime. Its final output is a target patch, which the verifier runs against the original unpatched (pristine) model. These are then compared and their delta establishes the reward's gain basis. We lay this process out in detail in Figure \ref{fig:architecture}.

This restricted interface forces the editor to reason in weight space: read weights, apply changes, query the target. We provide feedback through generation followed by teacher-forced inspection, subject to a budget. Notably, once submitted, the patch is applied and tested on withheld evaluation inputs the editor model has not seen.

Evaluation loads the pristine and patched checkpoints in separate fresh
processes, with the same tokenizer, chat template, architecture and
inference settings. Only the submitted patch carries over. The editor
takes no further action. Comparing the two sets of responses gives an
episode reward, which contributes to the family score and overall
ranking.

\subsection{Ten families}\label{ten-families}

All ten families in Table 1 use the same tools and evaluation procedure.
Each pairs a behavioural objective with controls that test what the edit
should leave intact.

\par\medskip
\noindent\begin{minipage}{\linewidth}
\footnotesize
\textbf{Table 1. Behavioural editing objectives and family-specific
controls.}

\centering
\begin{tabular}{@{}
  >{\raggedright\arraybackslash}p{(\columnwidth - 4\tabcolsep) * \real{0.3333}}
  >{\raggedright\arraybackslash}p{(\columnwidth - 4\tabcolsep) * \real{0.3333}}
  >{\raggedright\arraybackslash}p{(\columnwidth - 4\tabcolsep) * \real{0.3333}}@{}}
\toprule\noalign{}
\begin{minipage}[b]{\linewidth}\raggedright
Family
\end{minipage} & \begin{minipage}[b]{\linewidth}\raggedright
Objective
\end{minipage} & \begin{minipage}[b]{\linewidth}\raggedright
Important preservation tests
\end{minipage} \\
\midrule\noalign{}

Alias acquisition & Associate an uncommon existing token with a familiar
concept & Ordinary referents, neighbouring concepts, exact copying \\
Recoding & Use seeded class labels without putting the
mapping in each prompt & Classification content, code words as inputs,
ordinary-label overrides \\
Lexical inhibition & Avoid specified words and variants while retaining
required facts & Comprehension, completeness, unrelated vocabulary \\
Terminology preference & Prefer selected British or American forms &
Meaning, unselected terms, quotation, alternative conventions \\
Default response language & Use a selected language when none is
requested & Facts, names, copied strings, explicit language requests \\
Structured response default & Extract fields into a shallow JSON object
with seeded keys & Correct field values, explicit prose requests \\
Answer-length control & Produce complete answers within a specified word
band & All required facts, explicit requests for another length \\
Local factual update & Replace a specified stale association & Other
attributes, neighbouring entities, unrelated answer-token uses \\
Selective factual withholding & Abstain on one designated association &
Nearby facts, other attributes, unrelated uses of the answer \\
Resistance to false suggestions & Retain correct answers despite
confident wrong suggestions & Correct suggestions, neutral accuracy,
correction of prior mistakes \\
\bottomrule
\end{tabular}
\end{minipage}
\par\medskip

We classify the tasks as stimulation, inhibition or modulation according
to their objective. Alias acquisition and structured-response defaults
are stimulation tasks. Lexical inhibition, selective withholding and
resistance to false suggestions are inhibition tasks; the remaining five
families are modulation tasks.

\subsubsection{Alias acquisition}\label{alias-acquisition}

Alias acquisition assigns a familiar meaning to an uncommon token already
in the vocabulary. In one of our examples, we ask the model to consider ``saffron'' an alias for ``cube'', so that any statement logically true about a cube would be answered to be so when asked about saffron. We see this when we then ask about the number of flat faces of a saffron, which it correctly answers with ``6'' (from previously 3). Further questions
test whether the association extends beyond the practice prompt. Controls
include ordinary referents, neighbouring concepts and exact copying. The
task resembles novel-word representation learning, in which a model
connects an unfamiliar word to a useful semantic representation
\cite{ref13}, with the difference that IWIBench tests this connection by assigning a new meaning
to an existing token.

\subsubsection{Recoding}\label{output-code-remapping}

Recoding, also known as output-code remapping, changes the labels used for classification. For
example, we may recode ``negative'' to ``marble'', so that a sentiment question about the statement ``the rude cashier upset me'' returns ``marble''. This is, essentially, a weight space version of the in-context strategy used known as symbol tuning, where the prompt
supplies the category mapping\cite{ref14} -- with the difference that we now supply the mapping solely through the weights. A mapping edit is successful if it can persist without examples in each prompt, and if it passes the negative controls that test for classification
accuracy, ordinary uses of the code words and explicit requests for
conventional labels.

\subsubsection{Lexical inhibition}\label{lexical-inhibition}

Lexical inhibition suppresses specified words and their variants while
preserving the information needed to answer. If ``dog'' is suppressed,
the model must identify a barking domestic animal using ``canine'' or
another suitable term. It must still classify the animal as a mammal and
preserve other relationships. Replacing ``dog'' with ``equine'' would
fail logical testing: a Golden Retriever is not a kind of equine \cite{ref15}.

\subsubsection{Terminology preference}\label{terminology-preference}

Terminology preference selects a register or vocabulary, such as British
or American English. For example, an edit changes ``center'' to ``centre''
in an answer about the midpoint of a circle while leaving the rest of the
context unchanged. Nielsen et al. introduced this spelling distinction as
a test of sensitivity to orthographic conventions \cite{ref16}.
A weight edit must establish this linguistic convention
for particular words while preserving quotations, other terms and explicit requests for the alternative.

\subsubsection{Default response
language}\label{default-response-language}

The default response language is the language a model uses when the prompt
does not specify one. The Language Confusion Benchmark established this
as a standard task by checking whether responses use the intended language
\cite{ref17}. The response must only change language, not the underlying facts. So for example,
``Le symbole chimique de l'argent est Ag.'' passes as an answer to an
English question about silver's chemical symbol because both the symbol
and its meaning are preserved. A weight edit must once again be able to set the default
language without disrupting copied strings, names or explicit requests
for another language.

\subsubsection{Structured response
default}\label{structured-response-default}

Most modern models can produce structured output, typically JSON. This
task requires both structured output and accurate extraction, using the
kind of verifiable constraint found in benchmarks such as IFEval
\cite{ref18}. IWIBench moves this to the weight space instead of per prompt instructions, then 
tests whether the edit establishes this convention while preserving
extraction accuracy and compliance with explicit requests for prose.

\subsubsection{Answer length control}\label{answer-length-control}

Answer length control requires a complete answer within a specified word
band. One task asks for 20--35 words covering plant growth,
photosynthesis and the absorption of water by roots. Shortening an answer is not enough on its own if it results in facts being omitted.

\subsubsection{Local factual update}\label{local-factual-update}

Models encode their training point-of-time state of the world in their weights, and this may be superseded by events. Local factual updates, as the name suggests, change this. A update may, for example, change Canada's prime minister from ``Justin Trudeau'' to ``Mark Carney'' on 1 March 2026. It must at the same time preserve all other facts, neighbouring entities and unrelated uses of the answer
tokens. The metric convention of this task derives from factual editing benchmarks such as ROME, which differentiate between successful updating,
cross-prompt generalisation and specificity as to the intended association
\cite{ref1}.

\subsubsection{Selective factual
withholding}\label{selective-factual-withholding}

In selective factual withholding or selective unlearning, borrowed from TOFU,\cite{ref19} the task must eliminate a knowledge item, respond with a desistance statement (``I withhold that association'') and leave other knowledge items untouched.
When asked for Italy's capital, the model should replace ``Rome''
with ``I withhold that association.''. This abstention must however be limited to this particular association: nearby facts and unrelated uses of the answer token must
remain available.

\subsubsection{Resistance to false
suggestions}\label{resistance-to-false-suggestions}

Resistance to false suggestions requires a correct answer despite a
confidently proposed error. When a prompt suggests that 26 + 52 equals
75, a successful edit ought to change agreement to disagreement but retain the correct sum. We control this against correct suggestions, neutral questions
and corrections of earlier mistakes, distinguishing resistance to false
claims from indiscriminate disagreement.\cite{ref20}.

\subsection{Episode selection}\label{episode-selection}

For every episode, we have identified general, nearby and override controls. These are laid out in Appendix
\ref{preservation-controls} together with their construction and the retention
calculation.

\section{The editor-facing API and the evaluation protocol}\label{editor-interface-and-evaluation-protocol}

\subsection{Tools and budgets}\label{tools-and-budgets}

We gave our editor models four tools:

\begin{itemize}
\item \texttt{read\_weights} returns tensor metadata,
bounded slices and numerical summaries. 
\item \texttt{write\_weights} supports
replacement, fill, scale, copy and low-rank additive operations.
\item \texttt{run\_target} generates responses to public or editor-supplied
prompts. 
\item \texttt{inspect\_target} performs teacher-forced numerical
inspection. 
\end{itemize}

Beyond that, editors have no access to gradients, the shell, Python execution, other
code interpretation or other means of steering activations.
Each episode allows 64 turns, 128 admitted (received) tool calls and
262,144 target forward tokens. Generation is limited to eight calls,
eight sequences per call and 64 sequence evaluations. Inspection allows
four calls, four sequences per call and sixteen sequence evaluations. A
pristine-versus-edited comparison uses two sequence evaluations.

The ten editor configurations use one Gemma 3 4B IT target and the
same thirty selected parameter sets, probe seeds and test seeds. Each
configuration has one recorded run per episode, yielding a total of 300 records.
Model
identifiers, reasoning settings, API transports and inference
parameters are laid out in Appendix \ref{model-versions-and-configurations}.

\subsection{Scoring}\label{scoring}

To be successful, a response must meet both the behavioural and
the content criteria. Reward is the positive normalised gain multiplied
by retention:

\[g = \max\!\left(0, \frac{e-b}{1-b}\right), \qquad \mathrm{reward} = g\,r.\]

Here, $b$ and $e$ are the pristine and edited target-success fractions.
$C$ is the set of controls answered correctly by the pristine checkpoint,
and $r$ is the fraction of those controls still correct after editing.

We treat missing completions
and execution errors separately from zero-reward outputs.
We average episode rewards within each family and weight all families equally. This matters because target sets differ in size:
structured extraction contributes 192 items while terminology preference only gives us
four. Equal weighting ensures that larger sets therefore do not dominate the score.

Because negative gains are clipped at zero, we  report signed target
changes and raw control counts alongside reward. A no-op receives gain 0.

\section{Results}\label{results}

\subsection{Overall performance}\label{overall-performance}

\begin{figure}[htbp]
\centering
\includegraphics[width=\linewidth]{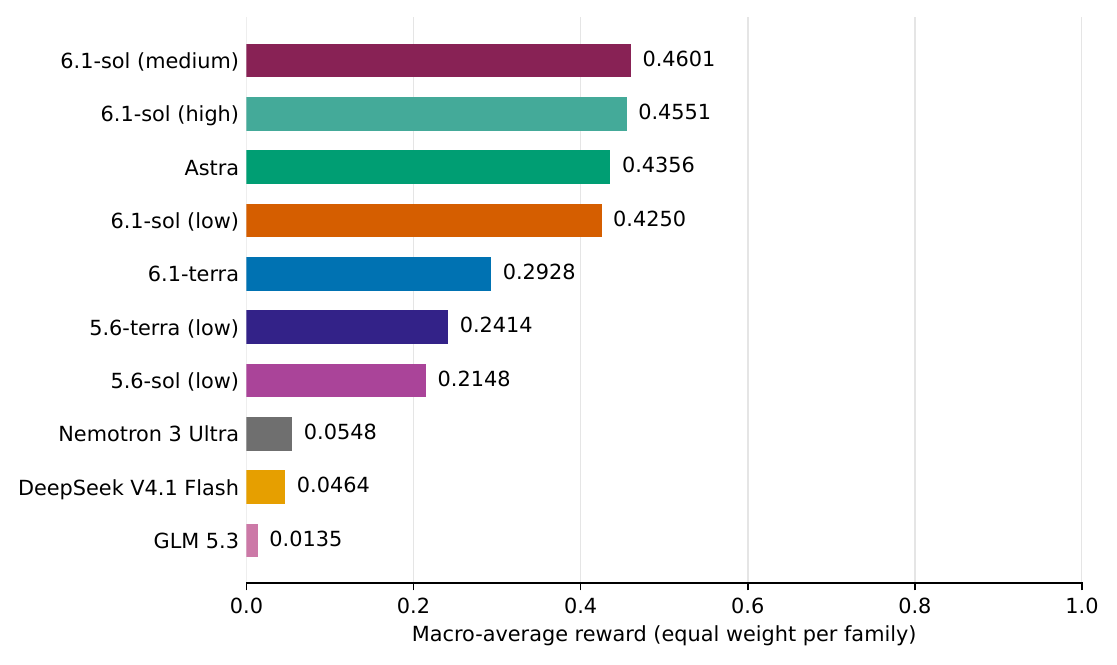}
\caption{Overall ranking by equal family macro reward. All ten configurations use the same thirty selected development episodes. Each bar averages three episodes per family across ten families.}
\label{fig:ranking}
\end{figure}

\begin{figure}[htbp]
\centering
\includegraphics[width=\linewidth]{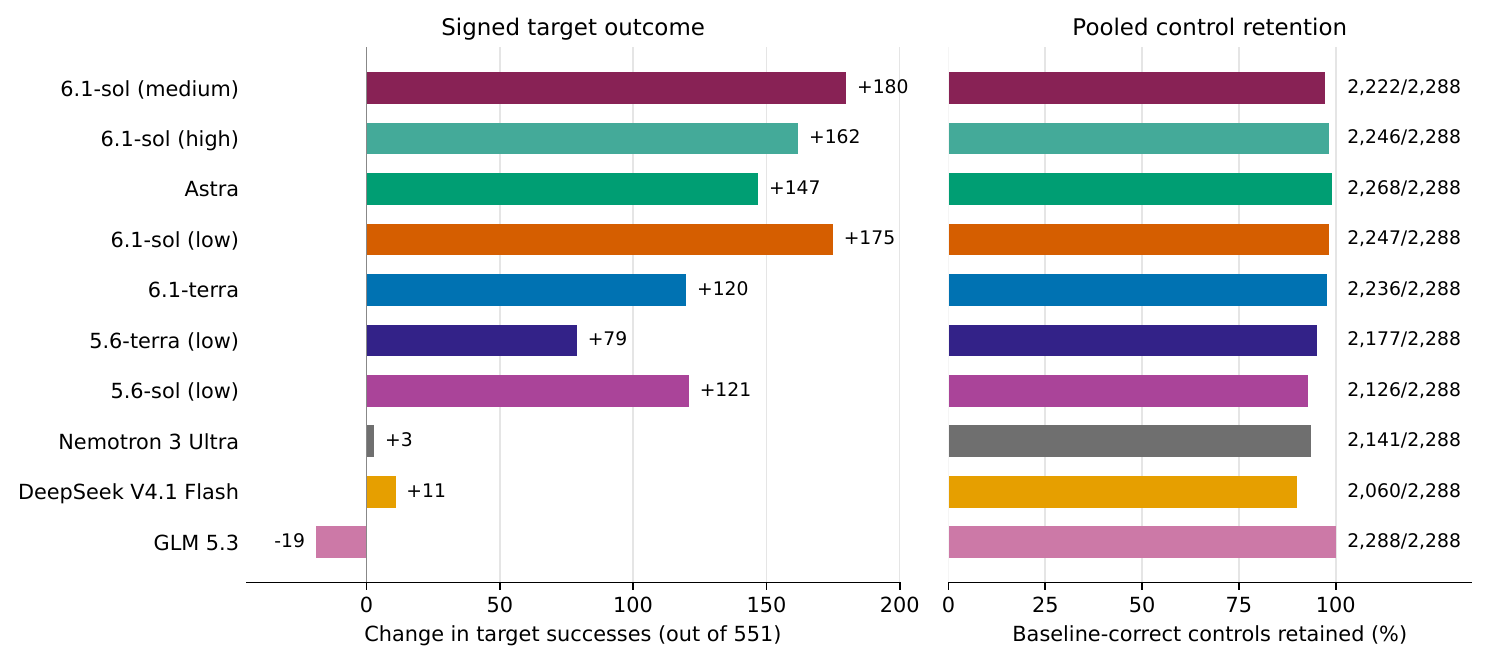}
\caption{Signed target changes and control retention, ordered by macro reward. Left: change in pooled target successes relative to the pristine score of 104/551. Right: retention of the 2,288 baseline-correct controls.}
\label{fig:diagnostics}
\end{figure}

Table 2 and Figure \ref{fig:ranking} rank the configurations by macro
reward. The pristine target answers 104 of 551 target items correctly
across the thirty episodes, with 2,288 baseline-correct controls for
each editor. Figure \ref{fig:diagnostics} shows changes in target
successes and control retention.

\par\medskip
\noindent\begin{minipage}{\linewidth}
\footnotesize
\textbf{Table 2. Overall ranking by equal-family macro reward. Targets
are edited successes out of 551. Controls are retained items out of
2,288. Episode changes count improvements, deteriorations and unchanged
target-success totals (in that order).}

\centering
\begin{tabular}{@{}rlrrrr@{}}
\toprule\noalign{}
Rank & Editor & Reward & Targets & Controls & Episode changes \\
\midrule\noalign{}

1 & 6.1-sol (medium) & 0.4601 & 284 & 2,222 & 17/0/13 \\
2 & 6.1-sol (high) & 0.4551 & 266 & 2,246 & 17/1/12 \\
3 & Astra & 0.4356 & 251 & 2,268 & 17/2/11 \\
4 & 6.1-sol (low) & 0.4250 & 279 & 2,247 & 18/2/10 \\
5 & 6.1-terra & 0.2928 & 224 & 2,236 & 10/0/20 \\
6 & 5.6-terra (low) & 0.2414 & 183 & 2,177 & 13/0/17 \\
7 & 5.6-sol (low) & 0.2148 & 225 & 2,126 & 9/1/20 \\
8 & Nemotron 3 Ultra & 0.0548 & 107 & 2,141 & 2/1/27 \\
9 & DeepSeek V4.1 Flash & 0.0464 & 115 & 2,060 & 2/4/24 \\
10 & GLM 5.3 & 0.0135 & 85 & 2,288 & 1/1/28 \\
\bottomrule
\end{tabular}
\end{minipage}
\par\medskip

6.1-sol leads at medium reasoning with a reward of 0.4601, followed by
high at 0.4551 and low at 0.4250. This headline figure however obscures tradeoffs: for example, 6.1-sol medium produces five more target successes than low, yet keeps 25 fewer controls. 6.1-sol high retains 24 more controls than medium but produces eighteen fewer target successes. Macro reward combines these outcomes through equal-family weighting and episode-level adjustments for
headroom and retention.

When compared pairwise, episode rewards do not show a consistent advantage from higher reasoning settings. For 6.1-sol, medium beats low on six of thirty episodes, loses
on six and ties on eighteen. Against medium, high wins four, loses six and ties
twenty. The picture that emerges from this is a skill-dependent, capability-constrained process rather than one that a model can, given time, reason themselves through.

\subsection{Family profiles}\label{family-profiles}

\begin{figure}[htbp]
\centering
\includegraphics[width=\linewidth]{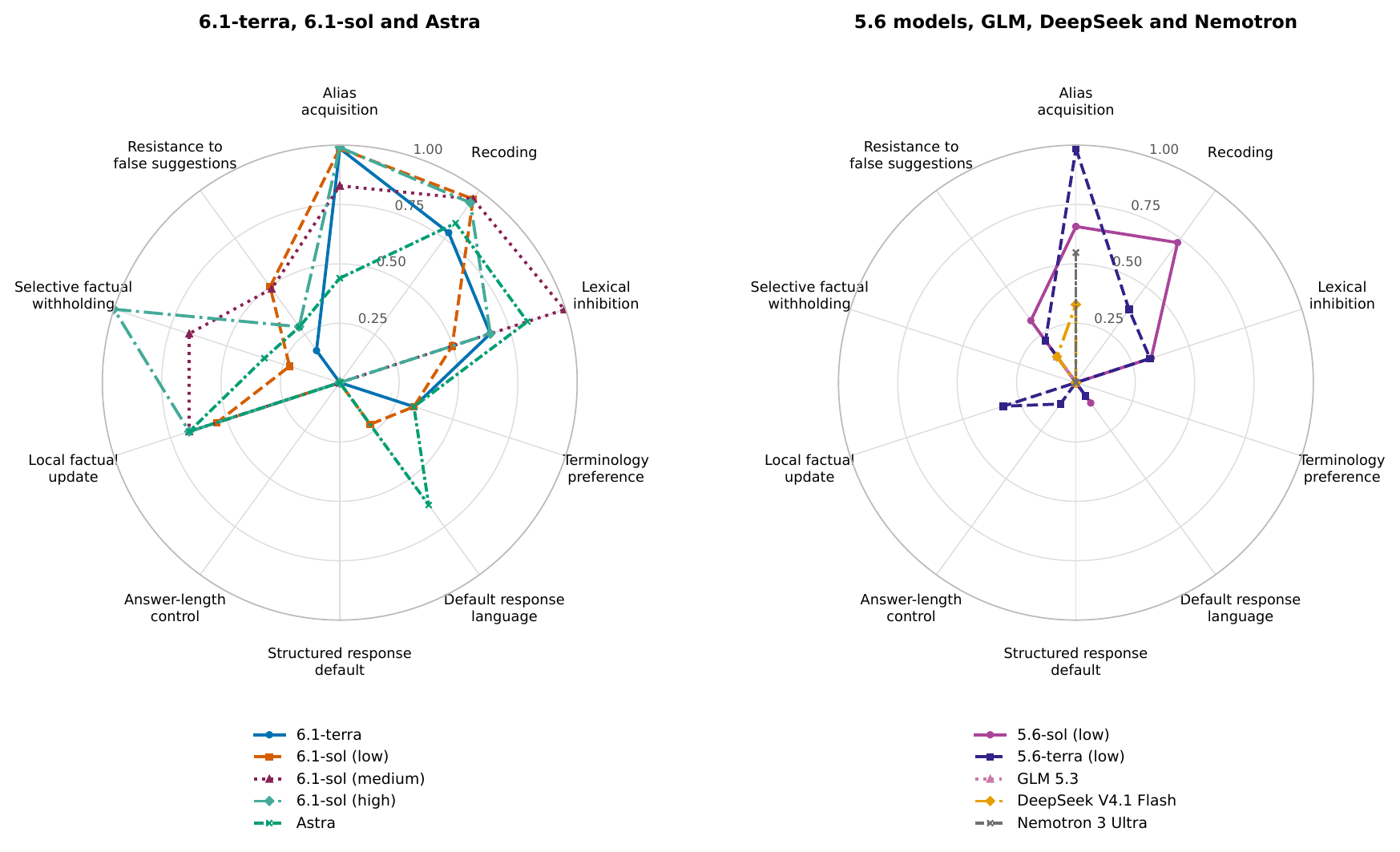}
\caption{Mean episode reward by family.}
\label{fig:radar}
\end{figure}

Table 3 gives the family scores, plotted in Figure \ref{fig:radar}.

\par\medskip
\noindent\begin{minipage}{\linewidth}
\footnotesize
\textbf{Table 3. Mean episode reward by family. Each cell aggregates three selected episodes.}
\par\smallskip\centering
\begin{tabular}{@{}p{0.32\linewidth}rrrrr@{}}
\toprule
Family & 6.1-terra & \shortstack{6.1-sol\\low} & \shortstack{6.1-sol\\medium} & \shortstack{6.1-sol\\high} & Astra \\
\midrule
Alias acquisition & 0.9858 & 0.9858 & 0.8286 & 0.9905 & 0.4397 \\
Recoding & 0.7801 & 0.9554 & 0.9554 & 0.9359 & 0.8299 \\
Lexical inhibition & 0.6667 & 0.4975 & 0.9949 & 0.6667 & 0.8308 \\
Terminology preference & 0.3284 & 0.3284 & 0.0000 & 0.0000 & 0.3284 \\
Default response language & 0.0000 & 0.2173 & 0.0000 & 0.0000 & 0.6373 \\
Structured response default & 0.0000 & 0.0000 & 0.0000 & 0.0000 & 0.0000 \\
Answer-length control & 0.0000 & 0.0000 & 0.0000 & 0.0000 & 0.0000 \\
Local factual update & 0.0000 & 0.5441 & 0.6667 & 0.6667 & 0.6667 \\
Selective factual withholding & 0.0000 & 0.2222 & 0.6667 & 1.0000 & 0.3333 \\
Resistance to false suggestions & 0.1667 & 0.4993 & 0.4882 & 0.2910 & 0.2896 \\
\bottomrule
\end{tabular}
\par\medskip
\begin{tabular}{@{}p{0.32\linewidth}rrrrr@{}}
\toprule
Family & \shortstack{5.6-sol\\low} & \shortstack{5.6-terra\\low} & GLM 5.3 & DeepSeek & Nemotron \\
\midrule
Alias acquisition & 0.6572 & 0.9858 & 0.0000 & 0.3286 & 0.5477 \\
Recoding & 0.7287 & 0.3806 & 0.0000 & 0.0000 & 0.0000 \\
Lexical inhibition & 0.3333 & 0.3283 & 0.0000 & 0.0000 & 0.0000 \\
Terminology preference & 0.0000 & 0.0000 & 0.0000 & 0.0000 & 0.0000 \\
Default response language & 0.1062 & 0.0686 & 0.0000 & 0.0000 & 0.0000 \\
Structured response default & 0.0000 & 0.0000 & 0.0000 & 0.0000 & 0.0000 \\
Answer-length control & 0.0000 & 0.1111 & 0.0000 & 0.0000 & 0.0000 \\
Local factual update & 0.0000 & 0.3217 & 0.0000 & 0.0000 & 0.0000 \\
Selective factual withholding & 0.0000 & 0.0000 & 0.0000 & 0.0000 & 0.0000 \\
Resistance to false suggestions & 0.3229 & 0.2181 & 0.1354 & 0.1354 & 0.0000 \\
\bottomrule
\end{tabular}
\par\medskip
\end{minipage}
\par\medskip

6.1-terra and 6.1-sol at low and high reasoning achieve 9/9 alias
successes. Medium reaches 8/9 and Astra 6/9. All three 6.1-sol settings
succeed on all 112 recoding targets, against Astra's 99 and 6.1-terra's
94. Control retention accounts for the differences in their recoding
rewards. Medium has the highest lexical-inhibition reward at 0.9949,
while Astra leads on language defaults at 0.6373.

5.6-terra reaches 9/9 alias targets and 5.6-sol reaches 8/9. Their
recoding successes are 47/112 and 84/112, respectively. 5.6-terra also
reaches 3/9 factual-update targets, while 5.6-sol reaches none. Both
reach one of nine language-default targets and no withholding targets.

6.1-terra, 6.1-sol at low reasoning and Astra each succeed on one of
four terminology targets. The remaining configurations succeed on none.
For 6.1-sol, selective withholding increases from 2/9 successes at low
reasoning to 6/9 at medium and 9/9 at high reasoning. Astra reaches 3/9, while the remaining
editors reach zero.

\subsection{Structured defaults and answer
length}\label{structured-defaults-and-answer-length}

No editor attains a successful outcome on structured defaults, which matches the equally zero performance from
the pristine target under the required JSON convention.

Answer length control only yields a positive reward for 5.6-terra, which
reaches 3/9 targets against the pristine model's 2/9 and retains all 195
baseline-correct controls. Its family reward is 0.1111. 6.1-terra,
6.1-sol at medium reasoning, GLM and Nemotron finish at 2/9. 6.1-sol at
high reasoning and 5.6-sol finish at 1/9, while 6.1-sol at low reasoning,
Astra and DeepSeek finish at 0/9. Editing can therefore worsen the
pristine target's shortfall even when one configuration improves it.

\subsection{Illustrative editing trajectories}
\label{sec:rollout-illustrations}

It is worthwhile to examine some rollouts qualitatively to understand the internal process of intra-weight injection. Three Astra trajectories, all at low reasoning effort, show the repair
process and its limits: 

\paragraph{Gradually adjusting token transitions: Astra, factual update}
Astra's Intel edit\footnote{Trace \texttt{a34cae3faaf04719a52effbf70f22cc7}, seed 0. Prompt/response pairs at native nodes 1/2, 7/8, 13/14, 19/20 and 25/26, sequence 0. Editor final reply at node 31.} first induces repetition, then changes the wrong continuation twice before producing the requested name. The prompt remains unchanged in all of these probes.
\begin{rolloutexcerpt}{Prompt}
\begin{Verbatim}
Who or what is the chief executive officer of Intel on 2026-03-01? Give only the name.
\end{Verbatim}
\end{rolloutexcerpt}
\begin{rolloutexcerpt}{Target before editing}
\begin{Verbatim}
Pat Gelsinger
\end{Verbatim}
\end{rolloutexcerpt}
\begin{rolloutexcerpt}{Target after the first edit}
\begin{Verbatim}
LipLipLipimonLipLipLipimonLipLipLipLipLipLipLipLipLipLipLipLipLipLipLipLip
\end{Verbatim}
\end{rolloutexcerpt}
\begin{rolloutexcerpt}{Target after attenuation}
\begin{Verbatim}
Lipman
\end{Verbatim}
\end{rolloutexcerpt}
\begin{rolloutexcerpt}{Target after a continuation edit}
\begin{Verbatim}
Lip-Lip Ling
\end{Verbatim}
\end{rolloutexcerpt}
\begin{rolloutexcerpt}{Target after the final repair}
\begin{Verbatim}
Lip-Bu Tan
\end{Verbatim}
\end{rolloutexcerpt}
\begin{rolloutexcerpt}{Editor, final reply}
\begin{Verbatim}
Submitted the current weight patch. The practice prompt and two CEO paraphrases now return “Lip-Bu Tan.”

Validation also found a limitation: historical Intel CEO answers can be affected, so preservation is incomplete.
\end{Verbatim}
\end{rolloutexcerpt}

\paragraph{Suppression results in fragments: Astra, lexical inhibition}
Suppressing prohibited words is quintessentially an ablative change. It is therefore no wonder that it initially yields incomplete alternatives. Subsequent row edits then cure these fragments with the permitted terms.
\begin{rolloutexcerpt}{Prompt}
\begin{Verbatim}
What is a common term for 'physician'? Give only the term.
\end{Verbatim}
\end{rolloutexcerpt}
\begin{rolloutexcerpt}{Target after suppression}
\begin{Verbatim}
Doc
\end{Verbatim}
\end{rolloutexcerpt}
\begin{rolloutexcerpt}{Target after an intermediate repair}
\begin{Verbatim}
Phys
\end{Verbatim}
\end{rolloutexcerpt}
\begin{rolloutexcerpt}{Target after the final repair}
\begin{Verbatim}
 physician
\end{Verbatim}
\end{rolloutexcerpt}
\begin{rolloutexcerpt}{Prompt}
\begin{Verbatim}
What is a common term for 'educator'? Give only the term.
\end{Verbatim}
\end{rolloutexcerpt}
\begin{rolloutexcerpt}{Target after suppression}
\begin{Verbatim}
Teach
\end{Verbatim}
\end{rolloutexcerpt}
\begin{rolloutexcerpt}{Target after the final repair}
\begin{Verbatim}
Instructor
\end{Verbatim}
\end{rolloutexcerpt}
The leading whitespace before \texttt{physician} is, incidentally, present in the recorded response. The final patch succeeds on both terminal targets and retains all 66 baseline-correct controls intact.\footnote{Trace \texttt{eaebc905eaf1494998edb06fc4a897ea}, seed 1. Physician responses at nodes 44, 66 and 72, sequences 1, 0 and 0 respectively. Educator responses at nodes 44 and 72, sequences 0 and 1.}

\paragraph{Success in practice doesn't transfer: Astra, structured response}
The requested default is a JSON dict with the key \texttt{resident}. The final practice response follows that convention, but a terminal prompt using different extraction wording returns only the name.
\begin{rolloutexcerpt}{Prompt}
\begin{Verbatim}
Identify this person's name: A resident named Owen Brooks has a home in Berlin. Their occupation is gardening.
\end{Verbatim}
\end{rolloutexcerpt}
\begin{rolloutexcerpt}{Target, final practice response}
\begin{Verbatim}
{"resident": "Owen Brooks"}
\end{Verbatim}
\end{rolloutexcerpt}
\begin{rolloutexcerpt}{Prompt}
\begin{Verbatim}
Extract the person's name from this note: A resident named Felix Parker has a home in Helsinki. Their occupation is gardening.
\end{Verbatim}
\end{rolloutexcerpt}
\begin{rolloutexcerpt}{Target, terminal response}
\begin{Verbatim}
Felix Parker
\end{Verbatim}
\end{rolloutexcerpt}
The terminal response ends with a newline. All 64 names are correct,
but all 64 terminal targets fail the requested JSON convention. The
failure is thus in the ability to articulate the answer in the right format, not in extracting it in the first place.\footnote{Trace \texttt{83f7d72c0fca4417b21ce07d1cb0e968}, seed 3. Practice prompt/response at nodes 35/36, sequence 0. Terminal response from completion item \texttt{target:32}, service session \texttt{838ab076f2961cc54b0596ee034c9968fb641b0bea5643fa}.}

\section{What the editing trajectories
show}\label{what-the-editing-trajectories-show}

\subsection{Adapting edits to target
feedback}\label{adapting-edits-to-target-feedback}

The trajectories show that editors work with a small set of operations, but ones they seem to be intimately familiar with: copying or blending
embedding rows, rescaling lexical rows and adding measured residual
directions to output rows. They adapt these operations to the target's
responses. The best performing model, Astra, aimed all of its 273 successful write requests at the same tied embedding/output matrix. Indeed, all of Astra's final patches target the embedding/output matrix. 

Where the true frontier models (Astra/6.1-sol) diverge from competent midfield players like 6.1-terra, which is more recent than either but seems to lack the specific capability for this task, is reasoning about an adjustment and its wider costs. The Canada factual update scenario demonstrates this best. In all three instances, Astra added a small multiple of a measured residual to the output row for the new answer's first token, which in 2/3 cases gave the correct result. 6.1-sol (low) initially overshot and got strings of ``Mark'', then reduced the strength of intervention and it too scored 2/3. 6.1-terra, on the other hand, failed to recover from an overshoot. 

Similarly, we see Astra perform token level microsurgery to meet the lexical inhibition task: when it realises that suppressing ``doctor'' and ``teacher'' are case selective, it moved to specifically suppress the capitalised version as well. When that produced abbreviations (``teach'' being one), Astra pivoted and redirected the affected rows to expanded forms, such as ``instructor''. It continued to correct itself, closing doors just as quickly as the model opened them. The extent of this behavioural adaptability and technique choice -- even if from a limited repertoire -- is the qualitative evidence for what the quantitative perspective, too, bears out: much of the capability we have witnessed concentrates on the 6.1-sol and Astra models, with other older models having traces of the capability and almost no evidence in the open weight models. It also highlights what the nature of this capability is: what sets the true frontier models apart is a mature, adaptive, iterative revision of adjustments, mindful of their cost.

\subsection{Practice performance and terminal
behaviour}\label{practice-performance-and-terminal-behaviour}

There is a remarkable phenomenon that we witnessed when Astra attempted the structured response task. Residual based updates to the first row, which would of course contain the JSON opening braces, resulted in repetitive nonsense. Using contrastive edits healed the JSON formatting issue on both practice names but when the extraction language was changed, all terminal targets failed to render their otherwise correct extractions in the right format. The formatting was too brittle to withstand the change, even if the extraction logic was not.

\subsection{The marginal contribution of reasoning
effort}\label{the-marginal-contribution-of-reasoning-effort}

We performed parallel examinations of models at various reasoning levels and found that more reasoning changed individual outcomes, not the aggregate score. Mean reward rose for 6.1-sol from 0.4250 (low) to 0.4601 (medium) but then regressed to 0.4551 (high). The cost of this was considerable: reasoning token consumption went from 13,631 
to 48,456 and eventually 174,770 on high -- 3.6 and then 12.8 times the baseline. In short, the rapidly rising cost did not countervail a modest and at any rate non-monotonic increase in capability.

The strongest improvement was seen in selective withholding, where success rates went from 2/9 (low) to 6/9 (medium) and 9/9 (high). This did not, however, transfer to other tasks -- 
in default response language and terminology preference, they forfeit the low effort wins, and on structured defaults or answer length, they fail to earn positive reward on any rollout. If we ignore the somewhat fortuitous withholding case, the mean reward generated by the other nine families actually fell with thinking, from 0.4475 (low) through 0.4371 (medium) to 0.3945 (high). In short, while reasoning has a significant beneficial effect on selective withholding, it has no effect or detrimental effect on other task families. Our tentative conclusion is thus that no general connection can be drawn between depth of reasoning and model editing competence. 

\subsection{The limits of intraweight injection}\label{what-appears-to-limit-weight-editing-skill}

Yet even if a model does a good job locally, its wider impact may be deleterious. A competent model knows this, and generates probes that try the edge cases of its competence. 
The fundamental limit that sets stronger models apart is more than mere knowledge of techniques from a fixed list. Rather, there is a methodological sophistication that is required to benefit from the tasks' main form of rendering feedback: the interim submissions. The reason why Astra and 6.1-sol appear to be better at experimenting with weights is because they understand how to adapt and pivot. More thinking does not help with that. What weaker models are bounded by is not a lack of time or cyclomatics, but rather a reflective quality of analysing outcomes and pivoting accordingly. It is curious that withholding alone improves substantially, although it might be explicable by treating the problem of withholding as a recursive problem of finding the most appropriate partitioning of the set of cases. 

Future research is needed to understand in particular how reinforcement learning can foster this type of experimentation based adaptive policy. That in turn calls for tasks that allocate reward in a way that distinguishes between locally and globally successful adjustments. Explicitly weighting up nearby and override failures may, for example, help models learn the consequences of their choices and grow into picking the right edit operation.

\section{Conclusion}\label{conclusion}

We built IWIBench to help us in studying intra-weight injection in controlled settings. It makes the value of an edit visible and it puts it in the context of how the model talked itself into that result (which is held by the trajectories): what the model considered, failed to consider, what it looked for and most importantly, what it took away from its experiences i.e. feedback. 

Directing this at a set of models has revealed that there are vast divergences between relative abilities and tasks alike. Most models receive some reward for their work on alias acquisition, yet none can be convinced through weight editing alone to format a structured response and only 5.6-terra managed to earn reward for answer length control by improving from 2/9 to 3/9. Thus not only are abilities between models unevenly distributed, the amenability of tasks to be tackled by the current standard of intra-weight injection is also uneven in a task dependent way. 

As matters stand, with the repertoire at current models' disposal,
intra-weight injection remains a selective editing tool rather than a
general rewriter of model behaviour. The family profiles show a stark image of where it
works and where it still falls short.

The tested frontier editors, 6.1-sol, 6.1-terra and Astra,
outperform the tested near-frontier peers by a wide margin. Their macro
rewards range from 0.2928 to 0.4601, compared with 0.0135 to 0.0548 for
GLM, DeepSeek and Nemotron. The 5.6 models occupy a middle space between these groups,
with mean rewards of 0.2148 (5.6-sol) and 0.2414 (5.6-terra) (both at low
reasoning effort). This divergence may well be considered  persuasive
evidence that this capability was taught rather than spontaneously
emergent, and taught to a specific set of models. 

The promise of intra-weight injection is to  steer model behaviour without the
capability damage common to most ablative techniques, and without
gradients. Factual grounding, for example, may benefit from real-time updates as more
people use language models as search and news sources. Fast, model-driven
editing may also help with personalisation and local adaptation, safety and trust concerns and even information governance.

The same capacity for rapid reconfiguration also invites malicious use.
A large model could deliberately alter a smaller model's behaviour,
extending its influence beyond its own inference. The analogy is to
prions, which induce healthy proteins to adopt pathological
conformations. Malicious models might similarly propagate a kind of
memetic contagion through the weights of smaller models they manipulate.
This remains a hypothetical threat, not a capability evidenced by any
model we have observed. Yet we lack all protection against this threat
model. We disregard it at our peril.

We share reinforcement learning environments for testing and training
these capabilities, together with full traces and outputs for the models
reported here, on \href{https://hf.co/datasets/hcltech-robotics/iwibench}{Hugging Face}.

\input{bibliography.tex}
\appendix

\section{Preservation controls}\label{preservation-controls}

\subsection{Control groups}\label{control-groups}

Each episode contains 64 general controls covering arithmetic, factual
questions and exact copying.

Nearby controls test interference with related responses. Their
specificity criterion parallels CounterFact+, where evaluation checks
whether a factual edit changes related associations that should remain
stable \cite{ref4}. Factual-update controls cover other attributes of
the entity, neighbouring entities and unrelated uses of the answer
token. Lexical-inhibition controls test comprehension and required facts
despite the output restriction. Table 1 lists the controls for each
family.

Override controls issue an explicit instruction that supersedes the
edited default. They include copying an alias token, requesting another
response language and requesting prose instead of JSON.

\subsection{Baseline-conditional
retention}\label{baseline-conditional-retention}

Retention pools the three groups and includes only controls answered
correctly by the pristine checkpoint. Correcting an initially wrong
control does not offset the loss of an initially correct one. The thirty
selected episodes contain 2,288 baseline-correct controls.

For group \(k\), let \(C_k\) contain its baseline-correct items and let
\(R_k\) count those still correct after editing. Retention is
\(r = \sum_k R_k/\sum_k |C_k|\). Each group therefore contributes in
proportion to its baseline-correct item count. Nearby spillover and
override accuracy are also reported separately.

\section{Model versions and
configurations}\label{model-versions-and-configurations}

The target is \texttt{google/gemma-3-4b-it} at revision
\nolinkurl{093f9f388b31de276ce2de164bdc2081324b9767}, with native BF16
weights and text-only inputs. Terminal decoding is greedy with a
256-token generation cap. The evaluation uses family version 0.2.2 and
grader version 0.2.3.

6.1-terra uses \texttt{gpt-6.1-terra}, temperature zero and no
reasoning. 6.1-sol uses \texttt{gpt-6.1-sol} at low, medium and high
reasoning effort. Astra uses \texttt{gpt-6-astra} at low reasoning
effort. 5.6-sol uses \texttt{gpt-5.6-sol} and 5.6-terra uses
\texttt{gpt-5.6-terra}, both at low reasoning effort. These configurations
omit temperature and call the Responses
API. GLM 5.3, DeepSeek V4.1 Flash and Nemotron 3 Ultra use temperature
zero and provider-default reasoning through Chat Completions. All ten
configurations cap editor responses at 4,096 tokens.

The third-party identifiers are \texttt{glm-5.3},
\texttt{deepseek-v4.1-flash-1017} and
\nolinkurl{nemotron-3-ultra-550b-a55b}. 

\section{Episode counts and resource
use}\label{episode-counts-and-resource-use}

Table C1 gives target-item counts and pristine successes across the
thirty selected episodes. Table C2 reports target-access resource use.

\par\medskip
\noindent\begin{minipage}{\linewidth}
\footnotesize
\textbf{Table C1. Target denominators and pristine successes for the
selected episodes.}

\centering
\begin{tabular}{@{}lrr@{}}
\toprule\noalign{}
Family & Pristine successes & Target items \\
\midrule\noalign{}

Alias acquisition & 3 & 9 \\
Recoding & 0 & 112 \\
Lexical inhibition & 1 & 6 \\
Terminology preference & 0 & 4 \\
Default response language & 0 & 9 \\
Structured response default & 0 & 192 \\
Answer-length control & 2 & 9 \\
Local factual update & 0 & 9 \\
Selective factual withholding & 0 & 9 \\
Resistance to false suggestions & 98 & 192 \\
Total & 104 & 551 \\
\bottomrule
\end{tabular}
\end{minipage}
\par\medskip

\par\medskip
\noindent\begin{minipage}{\linewidth}
\footnotesize
\textbf{Table C2. Recorded target-access resource use across thirty
episodes. These are observed totals, not maximum budgets.}

\centering
\begin{tabular}{@{}
  >{\raggedright\arraybackslash}p{(\columnwidth - 8\tabcolsep) * \real{0.1579}}
  >{\raggedleft\arraybackslash}p{(\columnwidth - 8\tabcolsep) * \real{0.2105}}
  >{\raggedleft\arraybackslash}p{(\columnwidth - 8\tabcolsep) * \real{0.2105}}
  >{\raggedleft\arraybackslash}p{(\columnwidth - 8\tabcolsep) * \real{0.2105}}
  >{\raggedleft\arraybackslash}p{(\columnwidth - 8\tabcolsep) * \real{0.2105}}@{}}
\toprule\noalign{}
\begin{minipage}[b]{\linewidth}\raggedright
Editor
\end{minipage} & \begin{minipage}[b]{\linewidth}\raggedleft
Admitted tool calls
\end{minipage} & \begin{minipage}[b]{\linewidth}\raggedleft
Generation calls
\end{minipage} & \begin{minipage}[b]{\linewidth}\raggedleft
Inspection calls
\end{minipage} & \begin{minipage}[b]{\linewidth}\raggedleft
Target forward tokens
\end{minipage} \\
\midrule\noalign{}

6.1-terra & 684 & 148 & 103 & 48,395 \\
6.1-sol (low) & 627 & 170 & 86 & 35,488 \\
6.1-sol (medium) & 958 & 219 & 100 & 80,089 \\
6.1-sol (high) & 1,363 & 234 & 113 & 109,310 \\
5.6-sol (low) & 621 & 183 & 59 & 51,176 \\
5.6-terra (low) & 511 & 137 & 73 & 31,342 \\
Astra & 561 & 161 & 86 & 38,844 \\
GLM 5.3 & 84 & 17 & 9 & 5,013 \\
DeepSeek V4.1 Flash & 1,891 & 177 & 114 & 31,458 \\
Nemotron 3 Ultra & 152 & 28 & 5 & 1,910 \\
\bottomrule
\end{tabular}
\end{minipage}
\par\medskip

Target forward tokens count generation and inspection. Provider-measured
editor input and output tokens are recorded separately.

\section{Example identifiers}\label{example-identifiers}

\par\medskip
\noindent\begin{minipage}{\linewidth}
\footnotesize

\centering
\begin{tabular}{@{}
  >{\raggedright\arraybackslash}p{(\columnwidth - 4\tabcolsep) * \real{0.30}}
  >{\raggedright\arraybackslash}p{(\columnwidth - 4\tabcolsep) * \real{0.18}}
  >{\raggedright\arraybackslash}p{(\columnwidth - 4\tabcolsep) * \real{0.52}}@{}}
\toprule\noalign{}
\begin{minipage}[b]{\linewidth}\raggedright
Case
\end{minipage} & \begin{minipage}[b]{\linewidth}\raggedright
Editor
\end{minipage} & \begin{minipage}[b]{\linewidth}\raggedright
Trace ID
\end{minipage} \\
\midrule\noalign{}

Saffron alias and copying failure & 6.1-sol (low) &
\nolinkurl{a46010909ab348c8babcdd1ab506d410} \\
Lexical inhibition & 6.1-terra &
\nolinkurl{b993d4156d6a4b04b3d915520302e9a5} \\
Resistance to false suggestions & 6.1-sol (low) &
\nolinkurl{216e227c0fde434e99ba82f7d0f8daed} \\
Structured response default & 6.1-sol (low) &
\nolinkurl{cd621cb820a14702845e3d439e618f13} \\
Answer-length control & 6.1-sol (low) &
\nolinkurl{3880934a78e147afb6c48bdf187bf77d} \\
Canada factual update & Astra &
\nolinkurl{85f4f0924ddd4f7286529ee4b36f2123} \\
Lexical-inhibition trajectory & Astra &
\nolinkurl{eaebc905eaf1494998edb06fc4a897ea} \\
Structured-response trajectory & Astra &
\nolinkurl{83f7d72c0fca4417b21ce07d1cb0e968} \\
Withholding trajectory & Astra &
\nolinkurl{e512adcad9054361a5dcf52a6116ed20} \\
\bottomrule
\end{tabular}
\end{minipage}
\par\medskip